\documentclass{article}
\usepackage{ijcai23}

\usepackage{times}
\usepackage{soul}
\usepackage{url}
\usepackage[hidelinks]{hyperref}
\usepackage[utf8]{inputenc}
\usepackage[small]{caption}
\usepackage{graphicx}
\usepackage{amsmath}
\usepackage{amsthm}
\usepackage{booktabs}
\usepackage{algorithm}
\usepackage{algorithmic}
\usepackage[switch]{lineno}
\usepackage{subcaption}
\usepackage{multirow}

\let\oldequation\equation
\let\oldendequation\endequation

\renewenvironment{equation}
  {\linenomathNonumbers\oldequation}
  {\oldendequation\endlinenomath}

\title{Sequential Recommendation with Probabilistic Logical Reasoning}

\author{
Huanhuan Yuan$^1$
\and
Pengpeng Zhao$^{1}$\thanks{ Corresponding authors.}\and
Xuefeng Xian$^{2*}$\And
Guanfeng Liu$^3$\And
Victor S. Sheng$^4$\And
Lei Zhao$^1$
\affiliations
$^1$Soochow University\\
$^2$Suzhou Vocational University\\
$^3$ Macquarie University \\
$^4$ Texas Tech University 
\emails
 hhyuan@stu.suda.edu.cn,
 ppzhao@suda.edu.cn,
xianxuefeng@jssvc.edu.cn,
guanfeng.liu@mq.edu.au,
 Victor.Sheng@ttu.edu,
 zhaol@suda.edu.cn
}

\begin{document}

\maketitle

\begin{abstract}
Deep learning and symbolic learning are two frequently employed methods in Sequential Recommendation (SR). Recent neural-symbolic SR models demonstrate their potential to enable SR to be equipped with concurrent perception and cognition capacities.
However, neural-symbolic SR remains a challenging problem due to open issues like representing users and items in logical reasoning.
In this paper, we combine the Deep Neural Network (DNN) SR models with logical reasoning and propose a general framework named \textbf{S}equential \textbf{R}ecommendation with \textbf{P}robabilistic \textbf{L}ogical \textbf{R}easoning (short for SR-PLR). 
This framework allows SR-PLR to benefit from both similarity matching and logical reasoning by disentangling feature embedding and logic embedding in the DNN and probabilistic logic network.
To better capture the uncertainty and evolution of user tastes, SR-PLR embeds users and items with a probabilistic method and conducts probabilistic logical reasoning on users' interaction patterns. 
Then the feature and logic representations learned from the DNN and logic network are concatenated to make the prediction.
Finally, experiments on various sequential recommendation models demonstrate the effectiveness of the SR-PLR. Our code is available at https://github.com/Huanhuaneryuan/SR-PLR.

\end{abstract}

\section{Introduction}
Sequential Recommendation (SR) has been proposed to solve information overload in a variety of real-world applications, including e-commerce, advertising, and other areas.
Meanwhile, Deep Neural Network (DNN) is widely used in SR to capture the sequential characteristics of user behaviors and generate accurate recommendations~\cite{hidasi_session-based_2016,kang_self-attentive_2018}. Recent developments in neural-symbolic methods~\cite{shi_neural_2020,ENRL,NS-ICF} have demonstrated competitive performance against DNN-based SR models, thus boosting significant research interest in combining DNN with symbolic learning.

Deep learning and symbolic learning are two different approaches frequently used in the field of artificial intelligence. 
The former is especially a central concern in current research with the resurgence data-driven learning paradigm. 
Without explicit common sense knowledge and cognitive reasoning, these data-hungry strategies are typically difficult to generalize~\cite{Marcus2020}. In contrast, symbolic learning involves the use of logical operators (e.g., AND ($\wedge$), OR ($\vee$) and NOT ($\neg$)) and human-interpretable representations of data (e.g., words or numbers) to perform language understanding or logical reasoning tasks.
As for SR, DNN-based works (such as SASRec~\cite{kang_self-attentive_2018}, etc.) sort to learning expressive representations of items and generating recommendations by calculating the similarity between the representations of historical interactions and target items. However, rather than only calculating the similarity score, symbolic learning-based models focus more on making predictions based on the users' cognitive reasoning procedure~\cite{chen_neural_2020}. For example, after buying a laptop, a user may prefer to purchase a keyboard rather than a similar laptop.

At the same time, DNN and symbolic learning are complementary, and fusing them properly could combine the strengths of both approaches, thus improving the performance of deep learning models~\cite{ENRL}.
For example, symbolic learning can provide a more flexible logical structure to latent features that are learned from DNN. Additionally, the introduction of deep learning enables end-to-end training of the symbolic learning and reasoning process. However, neural-symbolic learning for SR remains a challenging problem with open issues in the following aspects.
First, the most recent models for logical reasoning are embedding-based. The feature description and logical representation are coupled in the same framework, which makes it hard to distinguish which latent feature contributes to feature representation or logical reasoning~\cite{shi_neural_2020}. And intuitively, the representations that work for feature description and logical reasoning in the model may influence the recommendation differently. Second, most of them assume user preferences are static and embed users and items in a deterministic manner, but ignore that the user's tastes are full of uncertainty and evolving by nature, which incurs inaccurate recommendations~\cite{vae}.

In this paper, we enhance the classical DNN-based models with logical reasoning and propose a general framework named \textbf{S}equential \textbf{R}ecommendation with \textbf{P}robabilistic \textbf{L}ogical \textbf{R}easoning (short for SR-PLR). In our framework, feature embedding and logic embedding are disentangled in disparate networks, which enables SR-PLR to be benefited from both similarity matching and logical reasoning. Specifically, for the feature part, we take DNN-based SR methods (such as SASRec, GRU4Rec, etc.) as the backbone to learn the powerful latent representations of sequences. For the logical part, two transfer matrices are mentioned to convert the original feature embedding into several independent Beta distributions to represent historical items. Then two closed probabilistic logical operators (i.e., AND, NOT) are conducted on these distributions to infer the target items' distribution with the KL-divergence (Kullback-Leibler). Finally, the feature representation obtained from traditional SR methods is concatenated with the logical representation sampled from the output Beta distribution to make the prediction. 

In a nutshell, the contributions of our paper can be concluded as follows:
\begin{itemize}
    \item We propose a general framework for sequential recommendation that combines the deep learning method with symbolic learning.
    \item We develop a probabilistic embedding method for recommendation, and conduct probabilistic logical reasoning on users' interaction behaviors for better capturing the uncertainty and evolution of user tastes.
    \item We successfully implement our framework to traditional and newly released DNN SR models, and our experimental results show that the performance of all these models can be improved with the help of probabilistic logical reasoning.
\end{itemize}

\section{Related Work}
There are multiple topics related to our SR-PLR. Here, we first present some sequential recommendation works, and then introduce some neural-symbolic, probabilistic embedding recommendation models in this section.

\subsection{Sequential Recommendation}

In early works, Markov chains, Recurrent Neural Network (RNN), and Convolutional Neural Network (CNN) are commonly used in SR. For example, FPMC~\cite{FPMC} relies on modeling item-item transition relationships and predicts the next item with the last interaction of a user. GRU4Rec~\cite{hidasi_session-based_2016} captures the sequential patterns with a multi-layer Gate Recurrent Unit (GRU) structure and NARM~\cite{NARM} further imports the item-level attention mechanism into GRU to measure the importance of different items. Simultaneously,  Caser~\cite{caser} explores the application of CNN for sequential recommendation, which uses two types of convolutional filters to extract the information hidden in the users' sequences. Recently, SASRec~\cite{kang_self-attentive_2018} 
introduces self-attention into recommendation systems and achieves great success. Beyond that, MLP4Rec~\cite{ma2019hierarchical} simply stacks multiple MLP layers to model users’ dynamic preferences. S3Rec~\cite{zhou2020s3} and DuoRec~\cite{DuoRec} boost the performance with self-supervised signals.
Different from these models, we aim to integrate symbolic learning into these DNN models and endow SR with cognition ability.

\subsection{Neural-symbolic Recommendation}

Recent neural-symbolic recommendation models can be divided into two categories. One type of neural-symbolic model (e.g., ENRL~\cite{ENRL} and NS-ICF~\cite{NS-ICF}) aims to build an explainable recommender system with the aid of symbolic learning. They focus on learning interpretable recommendation rules on user and item attributes, and then output how these predictions are generated. Another type (e.g., LINN~\cite{shi_neural_2020}, NCR-E \cite{chen_neural_2020} and GCR~\cite{GCR}) explores to represent logical operators with the multilayer perceptron and conducts logical reasoning with the guide of several logical regularizers. In this way, sequential behavior can be presented as a logical expression, so that conducting logical reasoning and prediction in a continuous space. 
However, different from these methods to conduct reasoning with solely embedding, we disentangle feature and logic embedding in different networks, and then combine feature learning and symbolic learning in a unified framework.

\subsection{Probabilistic Embedding for Recommendation}
Probability distributions are widely used in representing
all the uncertain unobserved quantities in a model (including structural, parametric, and noise-related) and how they relate to the data~\cite{ghahramani2015probabilistic}. There are several works mentioned to model the users and items in probabilistic embeddings. For example, PMLAM~\cite{PMLAM} and DDN~\cite{DDN} represents users and items as learnable Gaussian distributions, and uses Wasserstein distance to estimate whether the user will buy the target item or not. DT4SR~\cite{DT4SR} learns the mean and covariance with different Transformers, which performs effectively for cold-start recommendation, and STOSA~\cite{STOSA} further proposes novel Wasserstein self-attention based on Gaussian distributions. Furthermore, some works use the deep generative model in SR with probabilistic methods. For example,  VSAN~\cite{vae} fuse Variational AutoEncoders (VAE) with self-attention networks to capture the long and local dependencies in the behavior sequences. Different from these above probabilistic models, we embed users and items as Beta distributions for not only capturing uncertainty but also facilitating logical reasoning.
\begin{figure*}[t]
\centering
\includegraphics[width=1\textwidth]{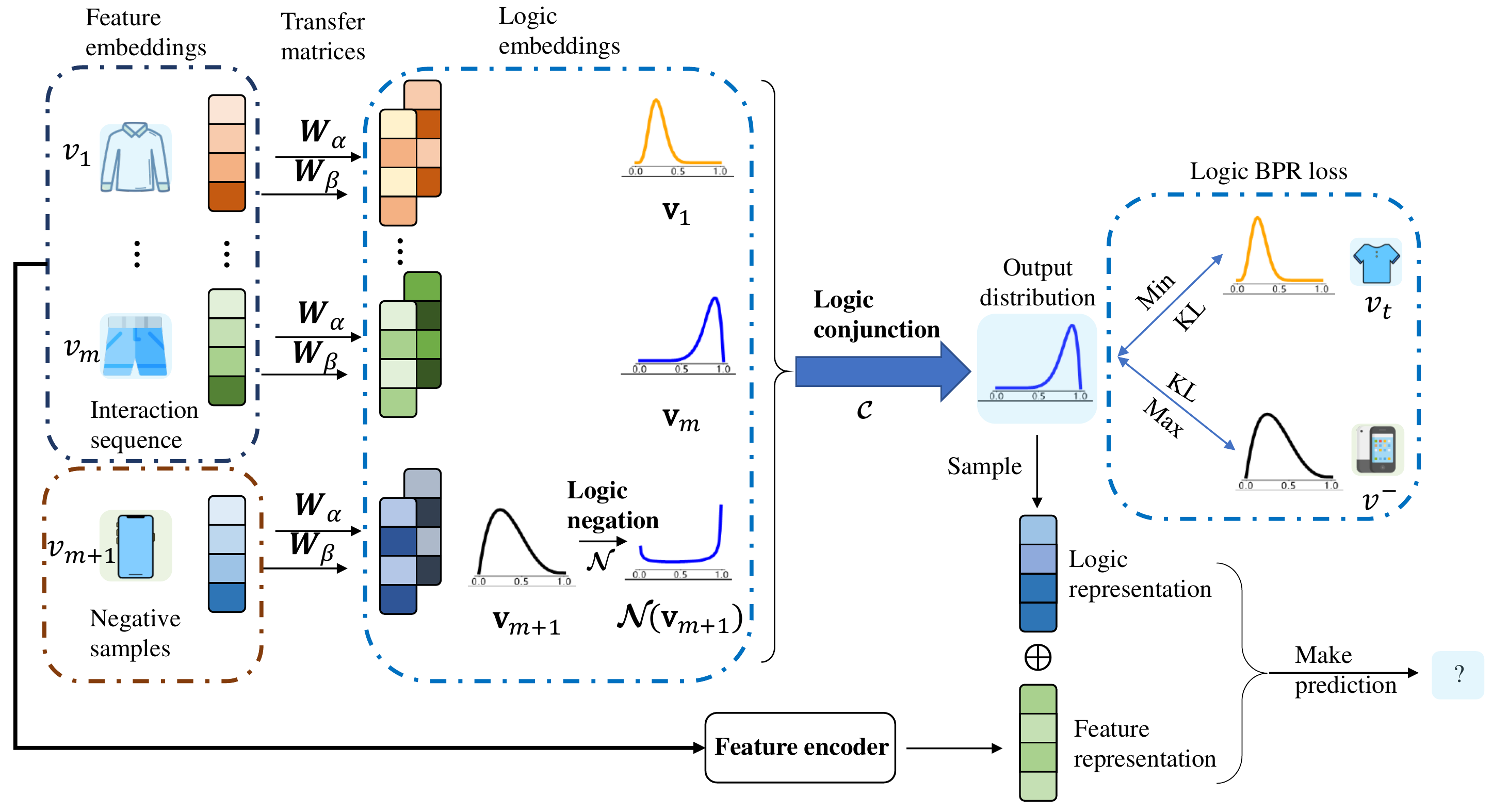}\\
\caption{The framework of SR-PLR. $v_{1}, v_{2}, \cdots, v_{m}$ are $u$'s historical items, $v_{t}$ is a target item and $\mathbf{v}_{i}$ is their corresponding distributions. $v_{m+1}$ and $v^{-}$ represent the sampled items that the user does not interact with.
}\label{framework}
\end{figure*}

\section{Formalization}

Formally, let $\mathbf{\mathcal{U}}$ and $\mathbf{\mathcal{V}}$ be user and item sets. Given $u$'s historical behaviors $ \mathbf{\mathcal{V}}_{u} = \{v_{1}, v_{2}, \cdots, v_{m}\} \subseteq \mathbf{\mathcal{V}}$, the goal of SRs is to predict the next item that $u\in \mathbf{\mathcal{U}}$ will interact with. 
For DNN SR models, they take $\mathbf{\mathcal{V}}_{u}$ as input to predict the most possible top-$K$ items, which can be formulated as 
$ v_{t}=\arg \max _{v_{i} \in \mathcal{V}} P\left(v_{m+1}=v_{i} \mid \mathbf{\mathcal{V}}_{u}\right)$,
where $v_{t}$ represents the target item of $u$.

In this paper, we complement classical DNN with logical reasoning.
For our logic part, the sequential recommendation task can be seen as a logical reasoning task. For example, let $\mathbf{\mathcal{V}}_{u}^{+}, \mathbf{\mathcal{V}}_{u}^{-}\subseteq \mathbf{\mathcal{V}}_{u}$ denote $u$'s clicked and unclicked item sets, $v_{1},v_{2} \in \mathbf{\mathcal{V}}_{u}^{+}$ and $v_{3}\in \mathbf{\mathcal{V}}_{u}^{-}$, the logical formula $v_{1}\wedge v_{2}\wedge (\neg v_{3})$ can be used to infer $v_{t}$.

\section{Methodology}

The SR-PLR framework shown as Figure~\ref{framework} consists of two main parts: feature representation learning and probabilistic logical reasoning. For the feature part, SR-PLR utilizes traditional DNNs (e.g., GRU4Rec, SASRec, etc.) as the feature encoder to extract the feature representations of users' behavior sequences. For the logic part, SR-PLR conducts logical reasoning with a 
probabilistic method, which will be fully explained from three points: probabilistic logic embedding, probabilistic logical operators and logical reasoning. More details are illustrated in the following parts.

\subsection{Feature Representation Learning}
We first briefly introduce how to learn feature representations in SR-PLR. 
Following the most widely used manner in SR models, the backbone feature encoder used in our framework contains two parts: the embedding layer and the feature encoder.
\subsubsection{Embedding Layer}
In our framework, ID information is all we need to build the model, since side information (e.g., sex, category, etc.) may be not always available in practice.
Hence, we embed the whole items' IDs into the same latent space~\cite{kang_self-attentive_2018} and generate the ID item embedding matrix $\mathbf{M} \in R^{|\mathbf{\mathcal{V}}| \times d}$, where $d$ is the embedding dimension. Given the $u$'s interaction sequence, the embedding of $\mathbf{\mathcal{V}}_{u}$ is 
initialized to $\mathbf{e}^{u} \in R^{m \times d}$ and $\mathbf{e}^{u}=\{\mathbf{m}_{1},\mathbf{m}_{2},...,\mathbf{m}_{m}\}$, where $\mathbf{m}_{k} \in R^{d}$ represents the item's embedding at the position $k$ in the sequence.

\subsubsection{Feature Encoder} Given the sequence embedding $\mathbf{e}^{u}$, a deep neural network model (e.g., SASRec) represented as $f_{\theta}(\cdot)$ is utilized to learn the representation of the sequence. The output representation of feature encoder $\mathbf{H}_{f}^{u} \in R^{d}$ is calculated as $\mathbf{H}_{f}^{u}=f_{\theta}(\mathbf{e}^{u})$, and chosen as the representation of $u$'s behavior sequence.

\subsection{Probabilistic Logical Reasoning}
As illustrated in the Introduction section, symbolic learning could grant DNN models the capability of cognition. 
However, how to represent users/items and how to conduct logical reasoning are still worth discussing. 

Hence in this paper, we: (1) design a probabilistic logical embedding method for the recommender system to represent users/items in the logical space, (2) apply probabilistic logical operators on these probabilistic embeddings, and (3) get the logical representations for the sequence by using logical reasoning. 

\subsubsection{Probabilistic Logical Embedding}
\label{emb}
Different from previous neural-symbolic works that treat the same ID embeddings as both feature representation and logical variable, two different types of embedding are used in SR-PLR to describe features and conduct logical reasoning, respectively. 

Specifically, we leverage the transfer matrix to covert ID embeddings into the logic space, and use multiple independent Beta distributions to demonstrate items thus modeling the uncertainty of user’s tastes. Where Beta distribution refers to a continuous probability distribution defined on [0, 1], and its Probability Density Function (PDF) is $p_{[(\alpha, \beta)]}(x) = \frac{1}{B(\alpha, \beta)}x^{\alpha-1}(1-x)^{\beta-1}$, where $x\in[0, 1]$, shape parameters $\alpha, \beta\in[0, \infty]$, and ${B(\alpha, \beta)}$ is the beta function.
Given ID embeddings, we use two transfer matrices represented as $\mathbf{W}_{\alpha}$ and $\mathbf{W}_{\beta} \in R^{d \times d}$ to transfer $\mathbf{M}$ into two shape matrices $\boldsymbol{\alpha}$ and $\boldsymbol{\beta}\in R^{m \times d}$:
\begin{equation}\label{trans}
\boldsymbol{\alpha} = \mathbf{M}\mathbf{W}_{\alpha},
\boldsymbol{\beta} = \mathbf{M}\mathbf{W}_{\beta}
\end{equation}
 For $v_{i}$, each dimension of $\mathbf{v}_{i}= {[(\boldsymbol{\alpha}_{i}, \boldsymbol{\beta}_{i})]}=[(\alpha_{i,1}, \beta_{i,1}), (\alpha_{i,2}, \beta_{i,2}), \cdots, (\alpha_{i,d}, \beta_{i,d})]$ characterizes the uncertainty by an independent Beta distribution instead of a single value, and its PDF $p_{\mathbf{v}_{i}}(x)$ is denoted as $\mathbf{P}(\mathbf{v}_{i}) = [p_{[(\alpha_{i,1}, \beta_{i,1})]}(x), p_{[(\alpha_{i,2}, \beta_{i,2})]}(x), \cdots, p_{[(\alpha_{i,d}, \beta_{i,d})]}(x)]$. Following BetaE~\cite{ren_beta_2020}, we clamp all elements in $\boldsymbol{\alpha}$ and $\boldsymbol{\beta}$ into (0, $10^9$] after multiplying transfer matrices. Note that items' representations in SR-PLR are assumed to follow the Beta distribution rather than the Gaussian distribution because we aim to guarantee the probabilistic logical operators on these Beta embeddings are closed. More details will be illustrated in the next section. 
\subsubsection{Probabilistic Logical Operators}
\label{ope}

In this section, we define two probabilistic operators on the Beta embedding space, which are called probabilistic negation operator ($\mathcal{N}$) and conjunction operator ($\mathcal{C}$), respectively. Note that since the disjunction operator can be implemented using negation and conjunction with De Morgan's laws~\cite{ren_beta_2020}, only the aforementioned two operators are discussed in our framework. 

\textbf{Probabilistic negation operator}:
For probabilistic negation operator, $\mathcal{N}(\mathbf{v}_{i})$ is defined as the reciprocal of $\mathbf{v}_{i}$:
\begin{equation}\label{not}
\mathcal{N}(\mathbf{v}_{i}) = {[(\frac{1}{\alpha_{i, 1}}, \frac{1}{\beta_{i, 1}}), (\frac{1}{\alpha_{i, 2}}, \frac{1}{\beta_{i, 2}}), \cdots, (\frac{1}{\alpha_{i, d}}, \frac{1}{\beta_{i, d}})]}
\end{equation}
It can be seen that, different from~\cite{shi_neural_2020} that relies on logical regularizations, $\mathcal{N}$ naturally satisfies $\mathcal{N}(\mathcal{N}(\mathbf{v}_{i})) = \mathbf{v}_{i}$. In addition, as shown in Figure~\ref{framework}, operator $\mathcal{N}$ can inherently reverse high probability density to low density and vice versa \cite{ren_beta_2020}, which enables $\mathcal{N}(\mathbf{v}_{i})$ to represent the dislikes item $v_{i}$ of users.


\textbf{Probabilistic conjunction operator}:
Given a user's behaviors $\mathbf{\mathcal{V}}_{u} = \{v_{1}, v_{2}, \cdots, v_{m}\}$ and their embeddings $\mathbf{v}_{1} = {[(\boldsymbol{\alpha}_{1}, \boldsymbol{\beta}_{1})]},\mathbf{v}_{2} = {[(\boldsymbol{\alpha}_{2}, \boldsymbol{\beta}_{2})]}, \cdots, \mathbf{v}_{m} = {[(\boldsymbol{\alpha}_{m}, \boldsymbol{\beta}_{m})]}$,
we define the output of conjunction $\overline{\mathbf{v}} = \mathcal{C}(\{ \mathbf{v}_{1}, \mathbf{v}_{2}, \cdots, \mathbf{v}_{m}\})$, where $\mathcal{C}$ is probabilistic conjunction operator. Following  \cite{ren_beta_2020}, $\overline{\mathbf{v}}$'s distributions can be represented as
\begin{equation}\label{and2}
\overline{\mathbf{v}} = [(\sum_{i=1}^{m}{\mathbf{w}_{i}\odot\boldsymbol{\alpha}_{i}}, \sum_{i=1}^{m}{\mathbf{w}_{i}\odot\boldsymbol{\beta}_{i}})]
\end{equation}
where $\sum$ and $\odot$ are the element-wise summation and product, respectively. $\mathbf{w}_{i} \in R^d $ is a weight vector. Its $j$-th dimension $w_{i,j}$ satisfy $\Sigma_{i=1}^{m}w_{i,j}=1$.
In this way, the PDF of $\overline{\mathbf{v}}$ is calculated as the weighted product of $\mathbf{\mathcal{V}}_{u}$' PDFs:
\begin{equation}\label{and1}
p_{\overline{\mathbf{v}}}(x) = \prod{p_{\mathbf{v}_{1}}^{\mathbf{w}_{1}}(x)p_{\mathbf{v}_{2}}^{\mathbf{w}_{2}}(x)\cdots p^{\mathbf{w}_{m}}_{\mathbf{v}_{m}}}(x)
\end{equation}
where $\prod$ is the element-wise product. 
In SR-PLR, we adopt the attention mechanism to learn the importance of different items during training:
\begin{equation}\label{att}
\mathbf{w}_{i} = \frac{\exp(MLP(\boldsymbol{\alpha}_{i} \oplus \boldsymbol{\beta}_{i}))} {\sum_{j}{\exp(MLP(\boldsymbol{\alpha}_{j} \oplus\boldsymbol{\beta}_{j}))}}
\end{equation}
where $\oplus$ represents concatenation operation, and $MLP: R^{2d}\rightarrow R^d$ is a multilayer perceptron taking the concatenation of $\boldsymbol{\alpha}$ and $\boldsymbol{\beta}$ as its input thus making our logical operators can be learned end-to-end. Obviously, the defined conjunction operator satisfies the equation $\mathcal{C}(\{\mathbf{v}_{1},\mathbf{v}_{1}, \cdots, \mathbf{v}_{1}\}) = \mathbf{v}_{1}$ and does not need any logical regularization.

It should be pointed out that both operators are closed in the Beta embedding space, which makes these operators could be combined in arbitrary ways and prevents exponential computation~\cite{ren_beta_2020}.

\subsubsection{Logical Reasoning}
\label{rea}
After defining logical embedding and operators, it is convenient for us to conduct reasoning on historical items. For example, given $v_{1},v_{2} \in \mathbf{\mathcal{V}}_{u}^{+}$ and $v_{3}\in \mathbf{\mathcal{V}}_{u}^{-}$, $v_{1}\wedge v_{2}\wedge (\neg v_{3})$ can be represented as $\mathcal{C}(\{\mathbf{v}_{1}, \mathbf{v}_{2}, \mathcal{N}(\mathbf{v}_{3})\})$ to calculate the target items' distribution.
Generally, for $v_{1}, v_{2}, \cdots, v_{m}\in \mathbf{\mathcal{V}}_{u}^{+}$ and sampled negative item $v_{m+1}, v_{m+2}, \cdots, v_{n}\in \mathbf{\mathcal{V}}_{u}^{-}$, we apply logical operators on them
\begin{equation}\label{left}
\overline{\mathbf{v}}_{u} = \mathcal{C}(\{\mathbf{v}_{1}, \mathbf{v}_{2}, \cdots, \mathbf{v}_{m}, \mathcal{N}(\mathbf{v}_{m+1}), \cdots, \mathcal{N}(\mathbf{v}_{n})\})
\end{equation}

As logical operators $\mathcal{N}$ and $\mathcal{C}$ are closed, $\overline{\mathbf{v}}_{u}$ is also a Beta embedding to represent the user's preference. Hence, by measuring the KL-divergence distance between $\overline{\mathbf{v}}_{u}$ and target item's distribution $\mathbf{v}_{t}$, the reasoning result of logic network can be obtained:
\begin{equation}\label{dist}
\mathrm{Dist}(\overline{\mathbf{v}}_{u}, \mathbf{v}_{t}) = \sum_{k=1}^{d}{\mathrm{KL}(\mathbf{P}_{k}(\mathbf{v}_{t}), \mathbf{P}_{k}(\overline{\mathbf{v}}_{u}))}
\end{equation}
where $\mathbf{P}_{k}(\overline{\mathbf{v}}_{u})$ represents the $k$-th dimension of $\mathbf{P}(\overline{\mathbf{v}}_{u})$. 
To this end, we utilize a pair-wise loss (e.g., Bayesian Personalized Ranking (BPR)~\cite{BPRMF}) to optimize the parameters of this logic network:
\begin{equation}\label{loss1}
\mathcal{L}_{l} ={\log(\sigma(\mathrm{Dist}(\overline{\mathbf{v}}_{u}, \mathbf{v}_{t})-\mathrm{Dist}(\overline{\mathbf{v}}_{u}, \mathbf{v}^{-}))}
\end{equation}
where $\mathbf{v}^{-}$ is a embedding of $v^{-}\in \mathbf{\mathcal{V}}_{u}^{-}$
, and $\sigma$ is the sigmoid function. 

\subsection{Training and Prediction}
To fuse DNN and logic network together, we are going to concatenate the feature representation $\mathbf{H}_{f}^{u}$ with logical representation $\mathbf{H}_{l}^{u}$ together to make the final prediction.
The most direct way to get logical representations is to take samples from the distribution $\overline{\mathbf{v}}_{u} =[(\overline{\boldsymbol{\alpha}}_{u},\overline{\boldsymbol{\beta}}_{u})]$. However, the sample operation is not differentiable, which makes the process hard to be trained end-to-end. Hence, 
in our framework, we choose the mean of distribution $\overline{\mathbf{v}}_{u}$ to represent the sequence:
\begin{equation}\label{sample}
\mathbf{H}_{l}^{u} = \frac{\boldsymbol{\overline{\alpha}}_{u}}{\overline{\boldsymbol{\alpha}}_{u}+\overline{\boldsymbol{\beta}}_{u}}
\end{equation}
Then, the corresponding prediction matrix
$\hat{\mathbf{y}}\in R^{|\mathbf{\mathcal{V}}|}$ can be generated by:
\begin{equation}\label{prey}
\hat{\mathbf{y}} = (\mathbf{H}_{f}^{u}\oplus\mathbf{H}_{l}^{u})(\mathbf{M} \oplus \mathbf{E})^{\top}
\end{equation}
where $\mathbf{E}= \frac{\boldsymbol{\alpha}}{\boldsymbol{\alpha}+\boldsymbol{\beta}}$.
And we use the cross-entropy loss to approximate the ground truth $\mathbf{y}$:
\begin{equation}
    \mathcal{L}_{Rec}=-\sum_{i=1}^{|\mathcal{V}|} y_{i} \log \left(\hat{y}_{i}\right)+\left(1-y_{i}\right) \log \left(1-\hat{y}_{i}\right)
\end{equation}
At last, the final objective function is:
\begin{equation}
    \mathcal{L}= \mathcal{L}_{Rec} + \lambda \mathcal{L}_{l}
    \label{LRec}
\end{equation}
where $\lambda$ is a hyperparameter.

\begin{table}
\centering
\scalebox{0.9}{
\begin{tabular}{lrrrrr}
\toprule
Datasets & Users & Items & Ratings & Avg. Len. & Sparsity\\
\midrule
Sports & 35,598 & 18,357 & 296,337 & 8.3 & 99.95\% \\
Toys & 19,413 & 11,925 & 167,597 & 8.6 & 99.93\% \\
Yelp & 30,499 & 20,068 & 317,182 & 10.4 & 99.95\% \\
\bottomrule
\end{tabular}}
\caption{Statistics of datasets.}
\label{tab:booktabs}
\end{table}
\begin{table*}[t]

\resizebox{0.77\width}{!}{
\renewcommand\arraystretch{1.25}{
\begin{tabular}{ccrrrrrrrrrrrr}
\hline
\multicolumn{2}{c}{\multirow{2}{*}{Models}} & \multicolumn{4}{c}{Amazon Sports}     & \multicolumn{4}{c}{Amazon Toys}     & \multicolumn{4}{c}{Yelp}     \\
\cline{3-14} 
\multicolumn{2}{c}{}  & \multicolumn{1}{c}{HIT@5} & \multicolumn{1}{c}{HIT@10} & \multicolumn{1}{c}{N@5} & \multicolumn{1}{c}{N@10} & \multicolumn{1}{c}{HIT@5} & \multicolumn{1}{c}{HIT@10} & \multicolumn{1}{c}{N@5} & \multicolumn{1}{c}{N@10} & \multicolumn{1}{c}{HIT@5} & \multicolumn{1}{c}{HIT@10} & \multicolumn{1}{c}{N@5} & \multicolumn{1}{c}{N@10} \\ \hline
\multirow{3}{*}{RNN}   & GRU4Rec   & 0.0186      & 0.0300    & 0.0121        & 0.0158         & 0.0352       & 0.0519        & 0.0240        & 0.0294         & 0.0217       & 0.0360        & 0.0144        & 0.0189         \\
    & GRU4Rec\_L   & \textbf{0.0225}       & \textbf{0.0353}        & \textbf{0.0141}        & \textbf{0.0182}         & \textbf{0.0387}       &\textbf{0.0557}         & \textbf{0.0268}        & \textbf{0.0323}         & \textbf{0.0249}       & \textbf{0.0433 }       & \textbf{0.0160}        & \textbf{0.0220} \\
    & Impro. & 20.97\% & 17.67\%  & 16.53\%  &15.19\%    & 9.94\% & 7.32\%  &11.67 \%  &  9.86\%  & 14.74\% & 23.06\%   & 11.11\%   & 16.40\%
    \\ \hline
\multirow{3}{*}{CNN}         & Caser        & 0.0141       & 0.0226        & 0.0087        & 0.0115         & 0.0183       & 0.0302        & 0.0113        & 0.0151         & 0.0231       & 0.0351        & 0.0164        & 0.0202         \\
    & Caser\_L     & \textbf{0.0173}       & \textbf{0.0283}        & \textbf{0.0109 }       & \textbf{0.0144}         & \textbf{0.0228}       & \textbf{0.0390}        & \textbf{0.0132}        & \textbf{0.0185}         & \textbf{0.0260}       & \textbf{0.0372}        & \textbf{0.0185}        & \textbf{0.0221} \\
    & Impro. &22.70\%  & 25.22\%  & 25.29\%  &25.27\%    & 24.59\% & 22.56\%  & 16.81\%  & 22.52 \%  & 12.55\% & 5.98\%   & 12.80\%   & 9.41\% \\ \hline
\multirow{3}{*}{Attention}   & SASRec       & 0.0317       & 0.0484        & 0.0172        & 0.0226         & 0.0630       & 0.0909        & 0.0354        & 0.0444         & 0.0422       & 0.0595        & 0.0322        & 0.0377         \\
    & SASRec\_L    & \textbf{0.0332}       & \textbf{0.0515}        & \textbf{0.0192}        & \textbf{0.0252}         & \textbf{0.0632}       & \textbf{0.0919}        & \textbf{0.0359}        & \textbf{0.0452}         & \textbf{0.0441}       & \textbf{0.0627}        & \textbf{0.0326}        & \textbf{0.0386}        \\    
    & Impro. &4.73\%  & 6.40\%  & 11.63\%  &11.50\%    &0.32 \% & 1.10\%  & 1.41\%  &  1.80\%  & 4.50\% & 5.38\%   & 1.24\%   & 2.39\% \\\hline
\multirow{3}{*}{DuoRec}      & DuoRec       & 0.0328       & 0.0505        & 0.0192        & 0.0249         & 0.0648       & 0.0929        & \textbf{0.0388}        & 0.0479         & 0.0434       & 0.0618        & 0.0319        & 0.0378         \\
    & DuoRec\_L    & \textbf{0.0342}      & \textbf{0.0522}        & \textbf{0.0200}        & \textbf{0.0257}         & \textbf{0.0652 }      & \textbf{0.0946}        & \textbf{0.0388}        & \textbf{0.0484}         & \textbf{0.0441}       & \textbf{0.0624}        & \textbf{0.0321}        & \textbf{0.0380}         \\
     & Impro. &4.27\%  & 3.37\%  & 4.17\%  &3.21\%    & 0.62\% & 1.83\%  & 0\%  &  1.04\%  & 1.61\% & 0.97\%   & 0.63\%   & 0.53\% \\\hline
\multirow{2}{*}{Logic}       & LINN         &0.0151  &0.0256   & 0.0101  &0.0132    &0.0196  & 0.0320  & 0.0133   & 0.0172    & 0.0215  &0.0355   &0.0146  &0.0192    \\
    & NCR-I        & 0.0162 & 0.0263   & 0.0110   &0.0146    & 0.0201  & 0.0322  &0.0135   &0.0176    &0.0204  &0.0354   &0.0146   &0.0191    \\ \hline
\end{tabular}
}
}
\caption{Overall performance on all datasets. `XX\_L' means the SR-PLR method that uses `XX' as the backbone and the numbers in bold indicate the better results that are generated by the same feature encoder. `N' denotes `NDCG' and `Impro.' denotes performance improvement compared with backbones.}
\label{overall}
\end{table*}
\begin{figure*}[t]
	\centering
	\subcaptionbox{Sports\label{a1}}{
		\includegraphics[width=5.5cm]{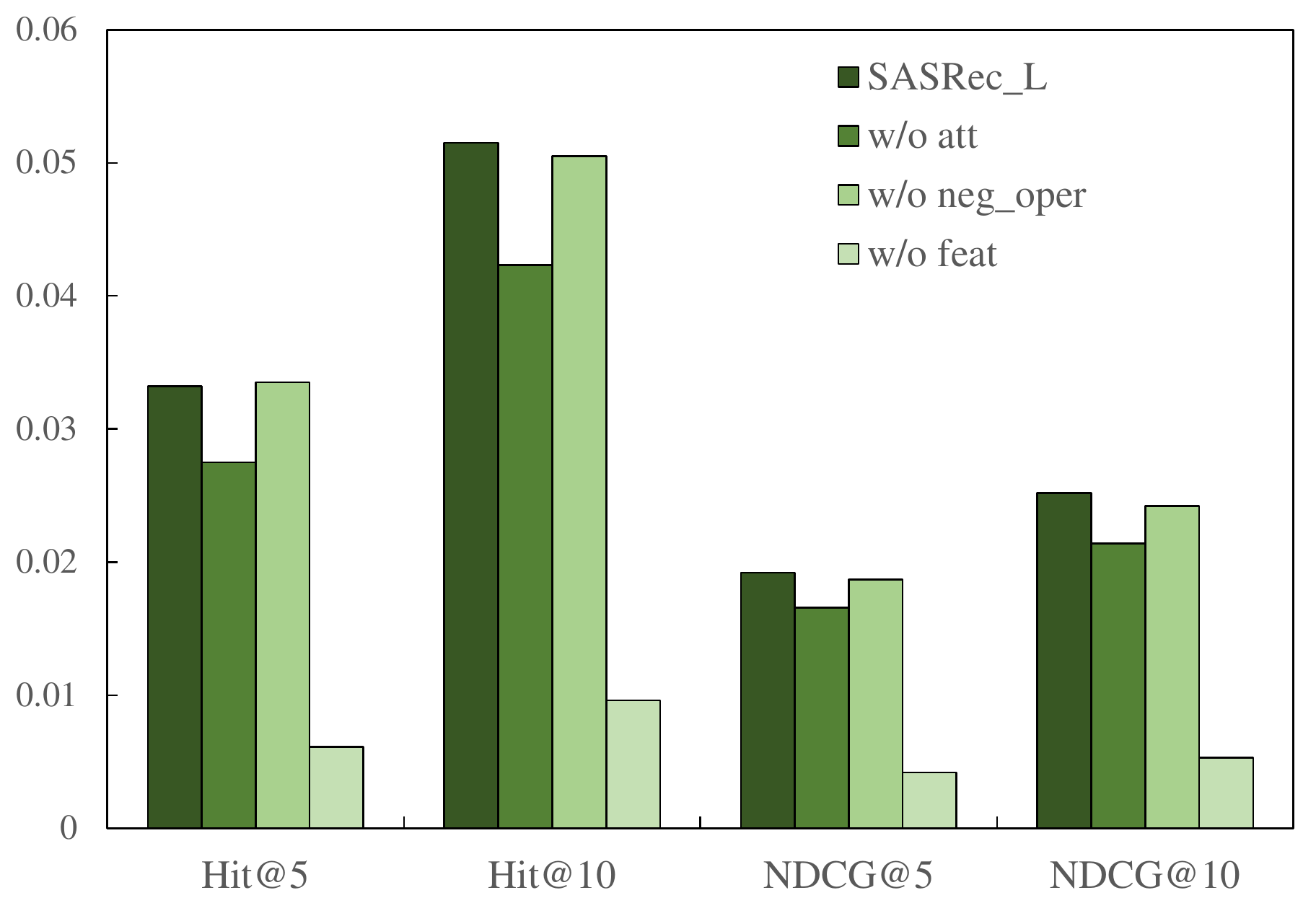}		
	}
	\hfill 
  \centering
	\subcaptionbox{Toys\label{a2}}{
		\includegraphics[width=5.5cm]{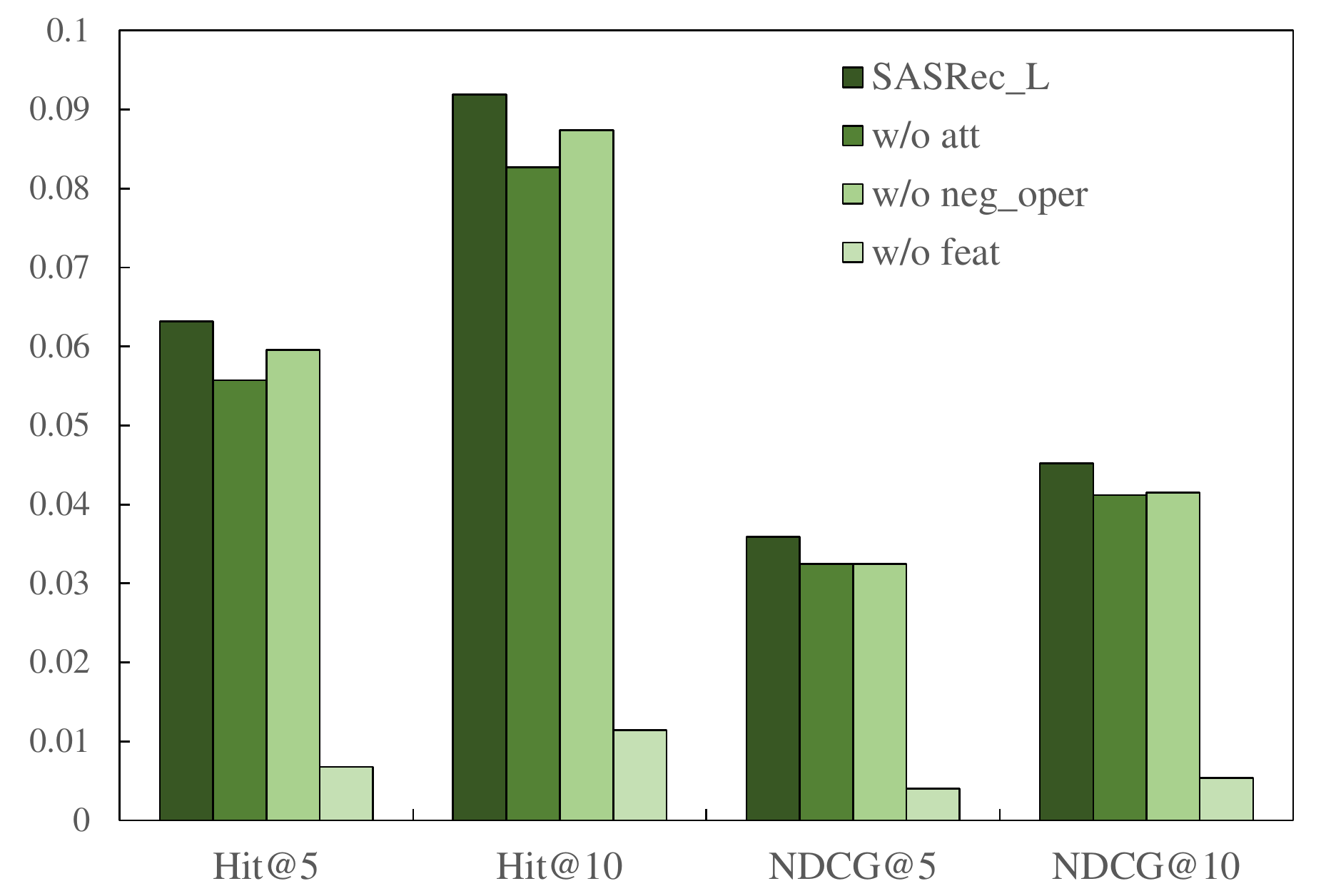}	
	}
 	\hfill 
   \centering
 	\subcaptionbox{Yelp\label{a3}}{
		\includegraphics[width=5.5cm]{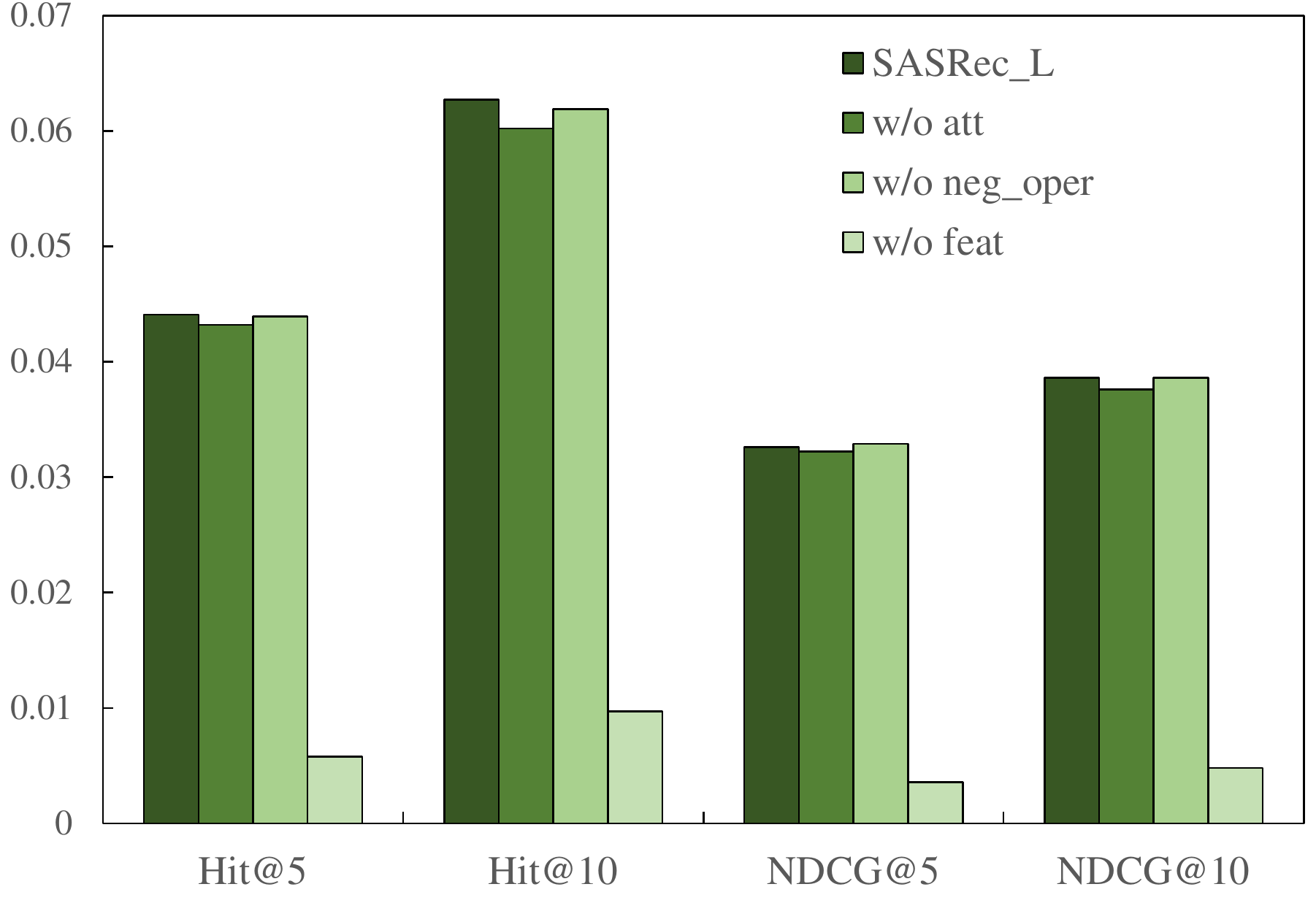}	
	}
	\caption{Ablation study of SR-PLR on three datasets.}
	\label{Ablation_Study}
\end{figure*}
\section{Experiments}

In this section, we conduct experiments with the aim of answering the following questions:
\textbf{Q1}: How do our models SR-PLR perform compared with other baselines?
\textbf{Q2}: What is the influence of key components of SR-PLR?
\textbf{Q3}: Whether is SR-PLR sensitive to the hyperparameters?
\textbf{Q4}: How is the robustness of SR-PLR?

\subsection{Experimental Settings}
\subsubsection{Dataset}
Experiments are conducted on three publicly available datasets. The statistics of each dataset are given in Table~\ref{tab:booktabs}.

\begin{itemize}
\item \textbf{Amazon}~\cite{Amazon}: \textit{Amazon Sports and Outdoors} and \textit{Amazon Toys and Games} are two sub-categories datasets crawled from Amazon, denoted as Sports and Toys.
\item \textbf{Yelp}: Yelp is one of the most widely used datasets for business recommendation.
\end{itemize}

Following previous works~\cite{kang_self-attentive_2018,caser}, we use the `5-core' version for all datasets and adopt the leave-one-out method to split these three datasets.

\subsubsection{Evaluation Metrics}

The widely used Normalized Discounted Cumulative Gain at rank $K$ (NDCG@K) and Hit ratio at rank $K$ (Hit@K) are used as evaluation metrics in our experiments, and we choose $K$ from {5, 10}. As recommended by~\cite{krichene_sampled_2020}, we adopt all-rank evaluation scores throughout the entire item set to ensure that the evaluation process is unbiased.
As~\cite{shi_neural_2020,chen_neural_2020} are evaluated with sampled metrics in the original paper, we compare SR-PLR with our re-implemented version.

\subsubsection{Baseline Methods}

Baselines used to compare with our models can be categorized into two groups. 

\textbf{DNN based models}:
For this group, we compare the Recurrent Neural Network (RNN), Convolutional Neural Network (CNN), self-attention based models, and newly released contrastive learning based model. They not only act roles as baselines, but also as the backbone of SR-PLR to demonstrate the effectiveness of combining logical reasoning.
\begin{itemize}
\item {\bf GRU4Rec}~\cite{hidasi_session-based_2016}: a method utilizing GRU to model user sequential behaviors as a strict order,
\item{\bf Caser}~\cite{caser}: a method using both horizontal and vertical convolution for sequential recommendation,
\item {\bf SASRec}~\cite{kang_self-attentive_2018}: a method based on the self-attention mechanism,
\item{\bf DuoRec}~\cite{DuoRec}: a method developing a  positive sampling strategy and using contrastive learning for SR with a model-level augmentation.
\end{itemize}
\textbf{Neural-symbolic models}:
For this group, we compare our framework with some neural-symbolic recommendation methods. As mentioned in~\cite{chen_neural_2020}, there are two different versions that base on implicit feedback NCR-I and explicit feedback NCR-E. Since SR-PLR is implicit feedback based, for fair comparisons, we use the implicit feedback versions here for these two neural-symbolic baselines.
\begin{itemize}
\item {\bf LINN}~\cite{shi_neural_2020}: a neural collaborative reasoning based recommendation method,
\item {\bf NCR-I}~\cite{chen_neural_2020}: a personalized method based on LINN.
\end{itemize}

\subsubsection{Implementation Details}
We run all methods in PyTorch~\cite{pytorch} with Adam \cite{Adam} optimizer on an NVIDIA Geforce 3070Ti GPU, and all these models are implemented based on RecBole~\cite{recbole}. 
The batch size and the dimension of embeddings $d$ are set to 2048 and 64 in our experiments. The max sequential length for all baselines is set as 50. We train all models 50 epochs. In the experiment, we keep all the hyperparameters of backbone models in RecBole unchanged and stack our logic network on them. SR-PLR is trained with a learning rate of 0.002. For the logic network, we set the $\lambda$ in Eq.~(\ref{LRec}) as a hyperparameter and select it from [0, 1] with step 0.1. For the negative item number in Eq.~(\ref{left}), we choose it from 1 to 10.

\subsection{Overall Performance (\textbf{Q1})}
We report the experimental results of different methods in Table~\ref{overall}. Note that `XX\_L' means the SR-PLR method that uses `XX' as the backbone. Where the `N' is short for NDCG and the numbers in bold indicate the better results that are generated by the same feature encoder. As we can see, the methods combining logical reasoning achieve better performance than other comparative models. Besides, there are more findings in these comparative experiments. 

Firstly, SASRec achieves better performance than GRU4Rec and Caser for both NDCG and Hit on all three datasets. It shows that modeling the importance of different interactions is highly effective for describing users' preferences. The main reason is that the items in the sequences contribute differently to final results, e.g., the laptop makes more contributions than clothes in buying the keyboard, which is also why the attention mechanism is used in our probabilistic operator.
Secondly, DuoRec performs better than SASRec in most cases, suggesting that pushing the positive views closer and pulling negative pairs away by contrastive learning are meaningful for getting better representations. 
Thirdly, both neural-symbolic models get similar results with GRU4Rec (consistent with the results reported in~\cite{shi_neural_2020,chen_neural_2020}). They also leverage deep neural networks to conduct logical reasoning, but only considering the logical equations of items may be not adequate for describing the complex relationship among interactions.

Finally, SR-PLR achieves performance improvement based on four types of feature encoders and performs best on three datasets. We contribute the improvement to the following aspects: (1) combining deep learning and symbolic learning in a dual feature-logic network, and (2) modeling users' dynamic preferences with a probabilistic method. As self-attention based SASRec performs better than RNN and CNN, meanwhile DuoRec is also the development of SASRec, we pay more attention to SASRec and perform more detailed experiments based on SASRec\_L.
\subsection{Ablation Study (\textbf{Q2})}

In this section, we conduct three ablation studies on all three datasets. We are going to examine the effectiveness of main three components, which are the attention mechanism in the probabilistic conjunction operator, the probabilistic negation operator and the feature network. Hence, three variants (which are named `w/o att', `w/o neg\_oper' and `w/o feat') are designed to compare against SASRec\_L, and the results are shown in Figure~\ref{Ablation_Study}. The details of these variants are as follows:

\textbf{w/o att}.
We first evaluate the necessity of item-level attention in our probabilistic conjunction operator. The variant 'w/o att' means to compute sequence distribution by aggregating the hidden states via average pooling operations in Eq.~(\ref{att}). From Figure~\ref{Ablation_Study}, we can see that our SASRec\_L far outperforms its variant without the attention operation on three datasets. It indicates that using a fixed weight matrix will be a bottleneck to modeling the dynamic evolving tastes of users. Our attention based logical operation can adaptively measure the importance score for each item, which helps to improve the model performance.

\textbf{w/o neg\_oper}.
The negation operator is also an essential part of SR-PLR. For w/o neg\_oper, we use positive items and conjunction operator to calculate the distribution of sequences, that is using $\overline{\mathbf{v}}_{u} = \mathcal{C}(\{\mathbf{v}_{1}, \mathbf{v}_{2}, \cdots, \mathbf{v}_{m}\})$ in Eq.~(\ref{left}). From Figure~\ref{Ablation_Study}, it can be seen that the recommendation performance can be booted with the help of operator $\mathcal{N}$. Note that there is nearly no work to consider to conclude negative items during representing the interaction sequences except neural-symbolic based models. Our experiments indicate that the negative operator is vital to make logical reasoning complete, and using the unclicked items with the negative operator in symbolic learning may be another way to leverage unlabeled data, which may be useful for the semi-supervised recommendation.

\textbf{w/o feat}.
We also evaluate the effectiveness of the feature network. For w/o feat, we alternatively make the prediction by using $\hat{\mathbf{y}} = \mathbf{H}_{l}^{u} \mathbf{E}^{\top}$ for Eq.~(\ref{prey}). As shown in Figure~\ref{Ablation_Study}, only using the probabilistic logic network will result in a performance drop, which indicates the importance of combing feature learning with symbolic learning, and the effectiveness of probabilistic logic network in boosting feature learning.
\begin{figure}[t]
	\centering
	\subcaptionbox{Sports\label{b1}}{
		\includegraphics[width=2.625cm]{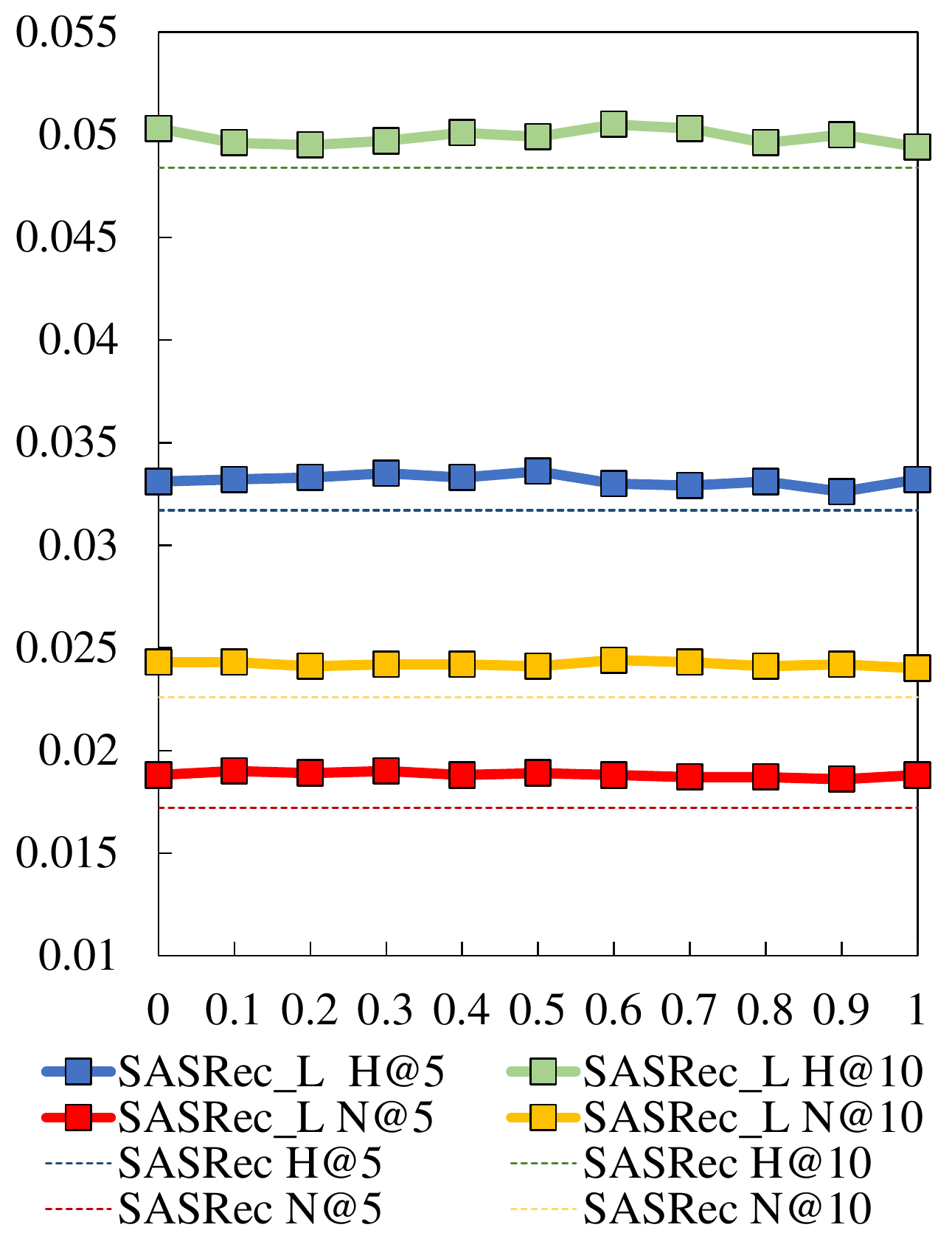}		
	}
	\hfill 
	\subcaptionbox{Toys\label{b2}}{
		\includegraphics[width=2.625cm]{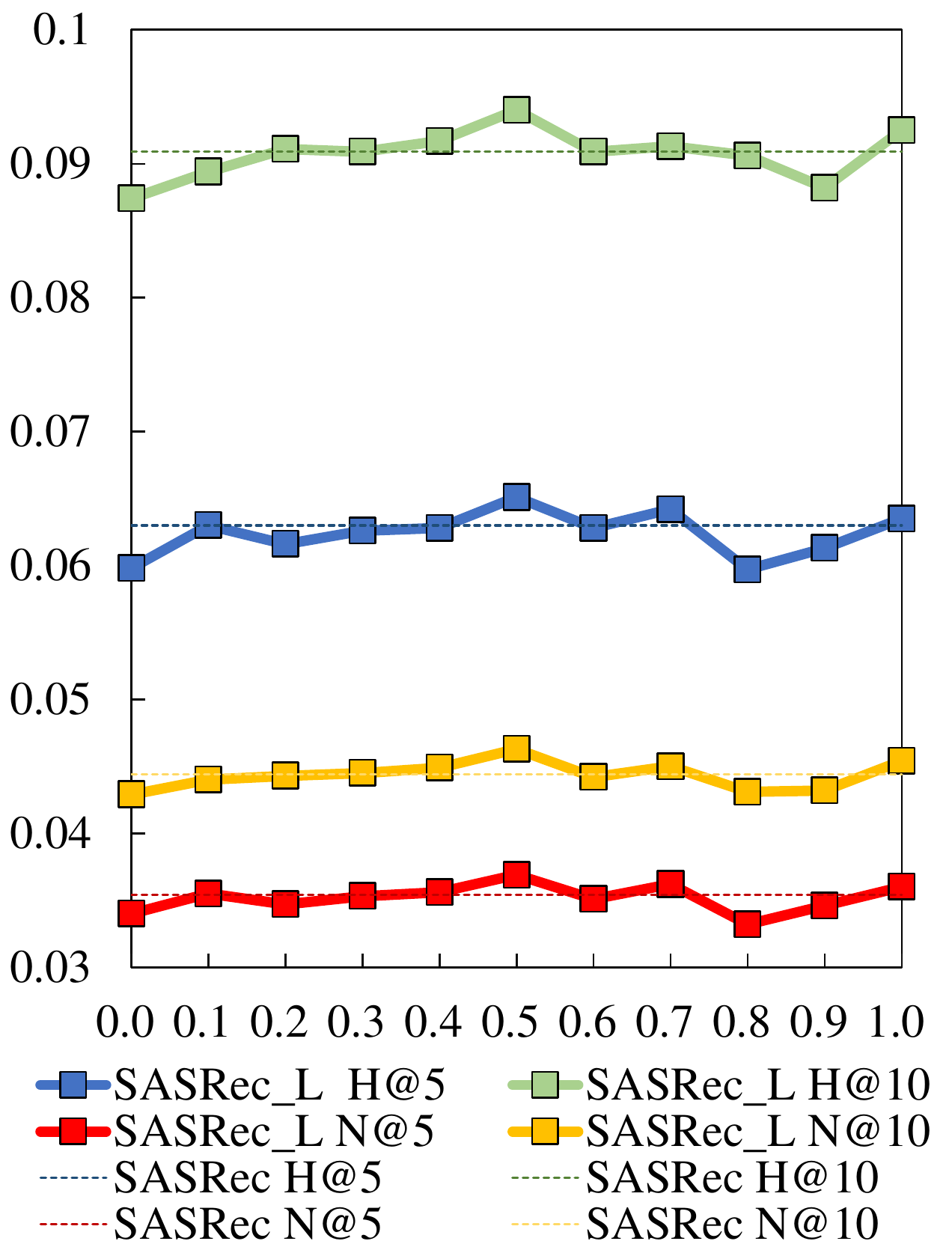}	
	}
 	\hfill 
 	\subcaptionbox{Yelp\label{b3}}{
		\includegraphics[width=2.625cm]{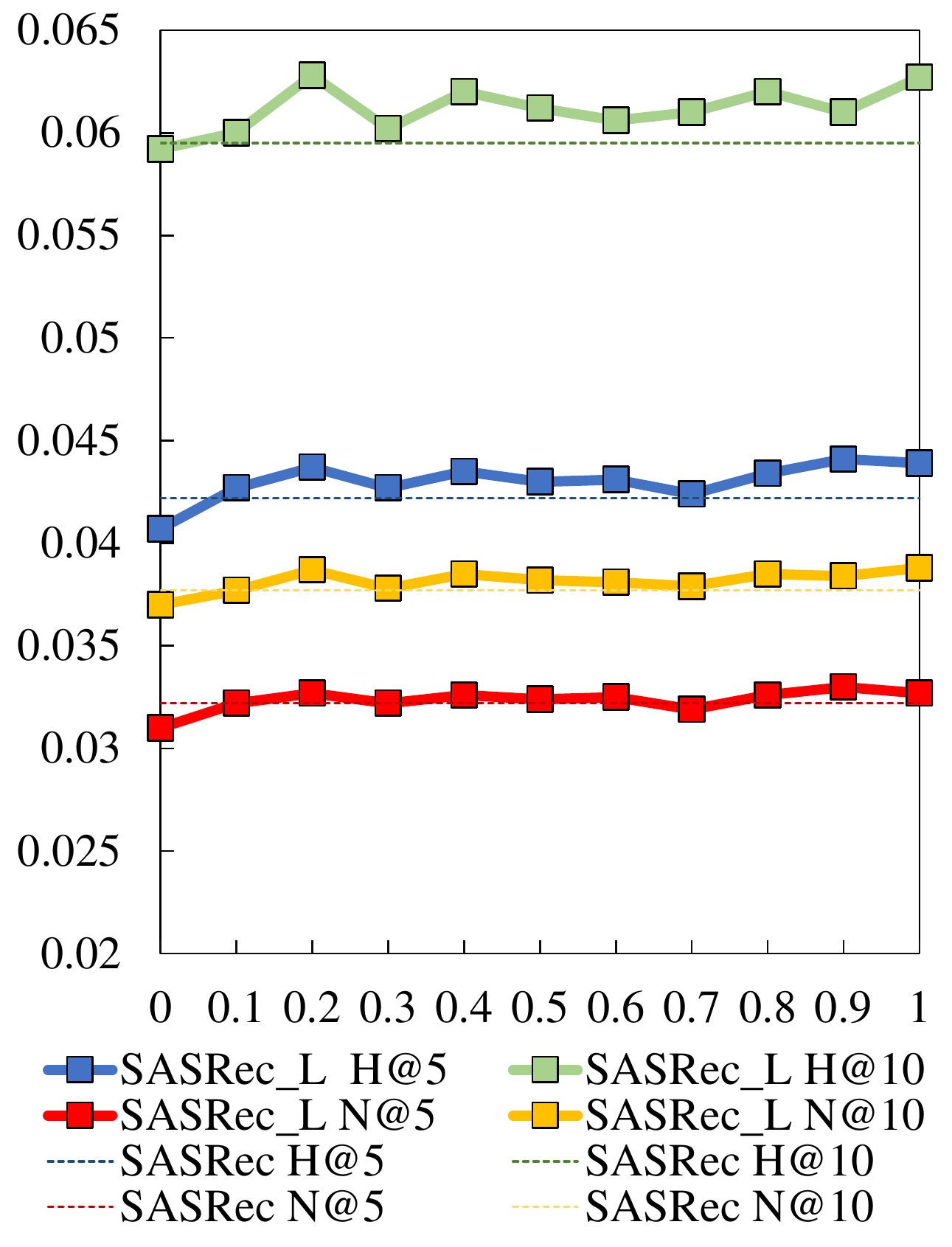}	
	}
	\caption{Sensitivity of $\lambda$ on three datasets.}
	\label{Sensitivity1}
\end{figure}
\begin{figure}[t]
	\centering
	\subcaptionbox{Sports\label{c1}}{
		\includegraphics[width=2.625cm]{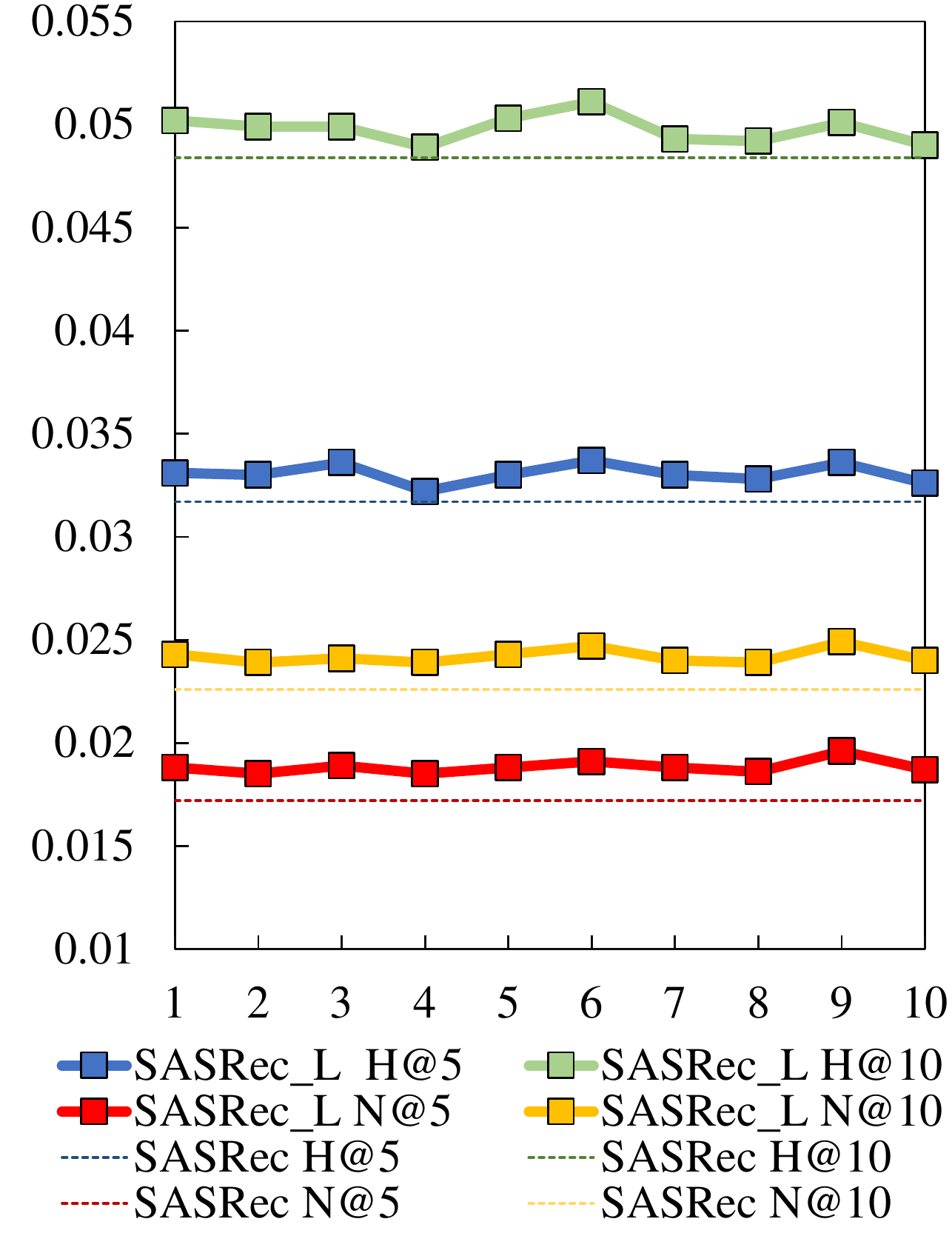}		
	}
	\hfill 
	\subcaptionbox{Toys\label{c2}}{
		\includegraphics[width=2.625cm]{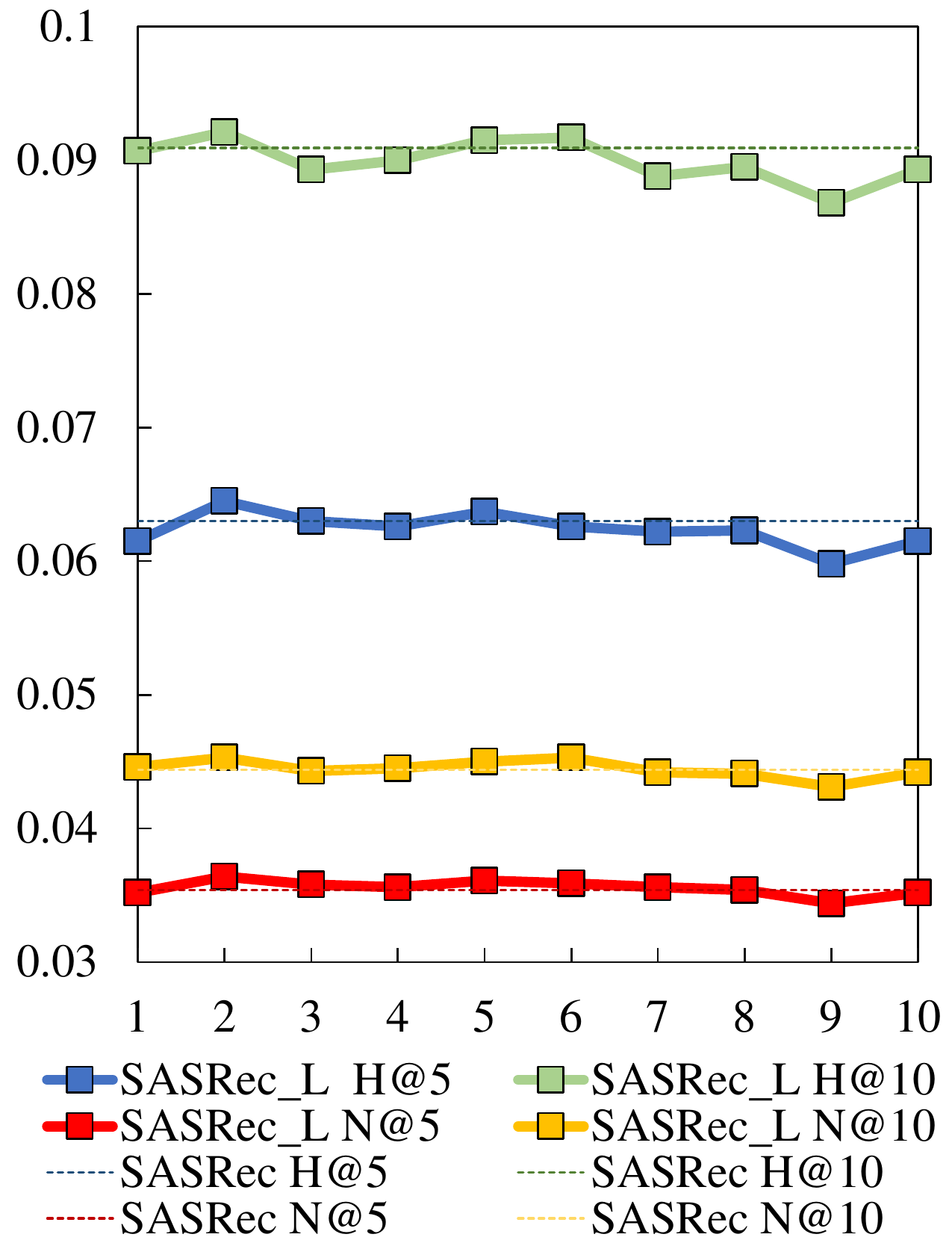}	
	}
 	\hfill 
 	\subcaptionbox{Yelp\label{c3}}{
		\includegraphics[width=2.625cm]{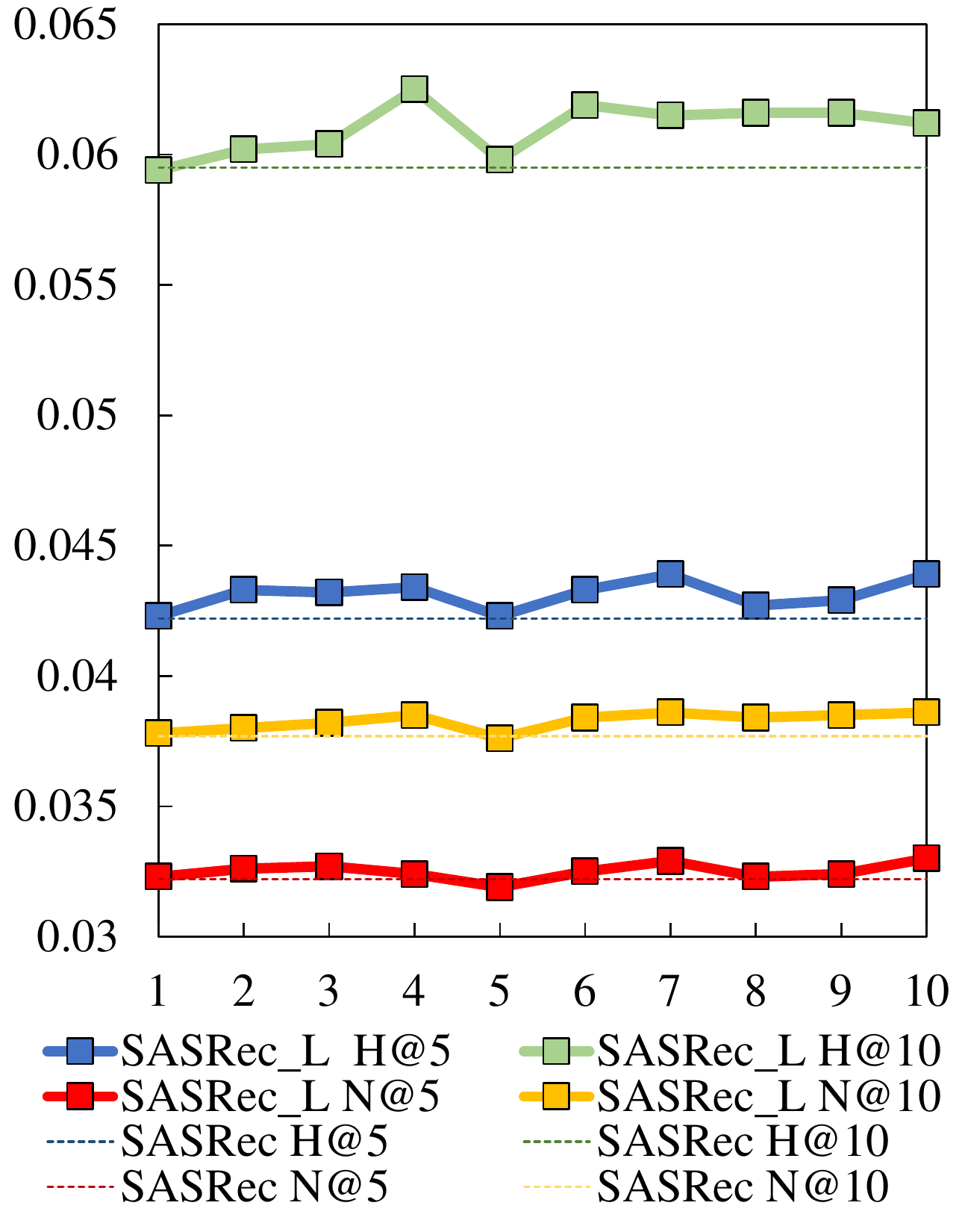}	
	}
	\caption{Sensitivity of negative item number on three datasets.}
	\label{Sensitivity2}
\end{figure}
\subsection{Hyper-parameter Sensitivity (\textbf{Q3})}
In this section, we are going to investigate how the performance varied with the changes of $\lambda$ and the sampled negative item number. In experiments, we vary one of these two hyperparameters and keep others unchanged to investigate the influence of $\lambda$ and the negative item number.

\textbf{Sensitivity of $\lambda$}.
For $\lambda$, we conduct the experiment in the range from 0 to 1 for all three datasets. The experimental results are shown in Figure~\ref{Sensitivity1}. In this figure, we compare SASRec\_L with SASRec under different $\lambda$. It can be seen that adding a logical regularizer, i.e., the $\mathcal{L}_l$ in Eq.~(\ref{loss1}), to optimize the logic network is helpful for the recommendation performance. Essentially, it acts as an auxiliary task to push the sequences' distribution close to the positive items' distributions and far away from the negative ones. In addition, it matters to carefully find a suitable value for each dataset, and values around 0.5 are recommended. 

\textbf{Sensitivity of sampled negative item number}.
For the negative item used for operator $\mathcal{N}$, we choose it ranges from 1 to 10, and the experimental results are indicated in Figure~\ref{Sensitivity2}.
It can be seen that, in most cases, introducing negative items is effective for logical reasoning, especially in Sports and Yelp. As we use implicit feedback in SR-PLR, the sampled negative items may be not truly disliked by the user, hence the number of negative items is set relatively small during training to maintain the semantic of sequence.  

\subsection{Robustness Analysis (\textbf{Q4})}
In the logic network, we endow SR-PLR with the capability of modeling uncertainty by using the Beta distribution to represent the items. To verify this point, we randomly mask some items in the sequence with probability $r$ during training and evaluate the performance of models in the original test dataset. Ideally, the item distribution contains more information than the static embedding, which makes SR-PLR could maintain its superiority with distorted sequences. As the results in Table~\ref{Roubust} illustrated, though both models' performances slowly drop, SASRec\_L still beats SASRec in most cases when $r$ varies from 0.1 to 0.5 in two datasets, which indicates SR-PLR is more robust by using the probabilistic method.

\begin{table}[t]
\center
\resizebox{0.78\width}{!}{
\setlength{\tabcolsep}{2pt}{
\begin{tabular}{@{}cccrrrrr@{}}
\toprule
\multicolumn{3}{c}{$r$}       & \multicolumn{1}{c}{0.1} & \multicolumn{1}{c}{0.2} & \multicolumn{1}{c}{0.3} & \multicolumn{1}{c}{0.4} & \multicolumn{1}{c}{0.5} \\ \midrule
\multirow{8}{*}{Sports} & \multirow{2}{*}{H@5} & SASRec &  0.0296   & 0.0292    & 0.0241    & 0.0214    & 0.0197    \\
 &        & SASRec\_L & \textbf{0.0334}    & \textbf{0.0310}    &  \textbf{0.0262}   &  \textbf{0.0215}   & \textbf{0.0198 }   \\ \cmidrule(l){2-8} 
  & \multirow{2}{*}{H@10} & SASRec &  0.0473   & 0.0462    & 0.0402    &\textbf{0.0361}     & \textbf{0.0327}    \\
 &        & SASRec\_L & \textbf{0.0499}    & \textbf{0.0484}   &  \textbf{0.0417 }  &  0.0357   & 0.0320    \\ \cmidrule(l){2-8}
  & \multirow{2}{*}{N@5} & SASRec &  0.0162   &  0.0158   &  0.0135   & 0.0122    &  0.0118   \\
 &        & SASRec\_L & \textbf{0.0189}    &  \textbf{0.0185}   & \textbf{0.0160}    &  \textbf{0.0136}   & \textbf{0.0124}    \\ \cmidrule(l){2-8}
 & \multirow{2}{*}{N@10}    & SASRec & 0.0219    & 0.0213    & 0.0187   & 0.0169    & 0.0160    \\
 &        & SASRec\_L & \textbf{0.0242}    &  \textbf{0.0241}   &  \textbf{0.0210}   &   \textbf{0.0182}  & \textbf{0.0163}    \\ \midrule
\multirow{8}{*}{Toys}   & \multirow{2}{*}{H@5} & SASRec &0.0582     &  0.0581   &  0.0504   & 0.0457    & 0.0384    \\
 &        & SASRec\_L & \textbf{0.0626}    & \textbf{0.0600 }   &\textbf{0.0582}     &  \textbf{0.0551}   &   \textbf{0.0446}  \\ \cmidrule(l){2-8} 
  & \multirow{2}{*}{H@10} & SASRec &0.0880     & 0.0872    &  0.0779   & 0.0700    & 0.0594    \\
 &        & SASRec\_L & \textbf{0.0890}    &\textbf{0.0888}     &\textbf{0.0879}     &  \textbf{0.0821}   &  \textbf{0.0733}   \\ \cmidrule(l){2-8}
  & \multirow{2}{*}{N@5} & SASRec &0.0329     &  0.0327   &  0.0285   & 0.0265    &  0.0228   \\
 &        & SASRec\_L & \textbf{0.0359 }   & \textbf{0.0343}    &\textbf{0.0337}     &  \textbf{0.0324}   &  \textbf{0.0271}  \\ \cmidrule(l){2-8}
 & \multirow{2}{*}{N@10}    & SASRec & 0.0426    & 0.0421    &  0.0374   &  0.0343   & 0.0295    \\
 &        & SASRec\_L & \textbf{0.0444}    & \textbf{0.0436}    &\textbf{0.0433}     &  \textbf{0.0411}   &  \textbf{0.0363}   \\ 
 \bottomrule
\end{tabular}
}
}
\caption{Robustness analysis on Sports and Toys.}
\label{Roubust}
\end{table}



\section{Conclusion}
In this paper, we proposed a general framework named SR-PLR to combine deep learning with symbolic learning via a Beta embedding method. Our main idea is two-fold: One is to disentangle items' embeddings and endow the basic DNN based sequential recommendation with cognitive capability; Another is to model the uncertainty and dynamic tastes of users. Therefore, we designed a dual feature-logic network and applied probabilistic logical operators on the items' Beta embeddings. Experiments on three real-world datasets showed that SR-PLR exhibits significant improvement against traditional baselines and neural-symbolic models.
Our future work will consider modeling content information and knowledge graph with more interpretable and transferable logic neural networks in sequential recommendation, and that may lead this framework to be more transparent and achieve significant performance.
\section*{Acknowledgments}
This research is partially supported by the NSFC (61876117, 62176175), the major project of natural science research in Universities of Jiangsu Province (21KJA520004), Suzhou Science and Technology Development Program (SYC2022139), the Priority Academic Program Development of Jiangsu Higher Education Institutions.

\bibliographystyle{named}
\bibliography{ijcai23}

\end{document}